\documentclass[acmsmall]{acmart}

\AtBeginDocument{%
 \providecommand\BibTeX{{%
 \normalfont B\kern-0.5em{\scshape i\kern-0.25em b}\kern-0.8em\TeX}}}

\setcopyright{none}
\renewcommand\footnotetextcopyrightpermission[1]{}
\usepackage{amsmath}
\usepackage{longtable}
\usepackage[utf8]{inputenc}
\usepackage[T1]{fontenc} 
\usepackage{hyperref} 
\hypersetup{
 colorlinks=true,
 linkcolor=blue,
 filecolor=magenta, 
 urlcolor=cyan
 }
\usepackage{url} 
\usepackage{booktabs} 
\usepackage{amsfonts} 
\usepackage{nicefrac} 
\usepackage{microtype} 
\usepackage{lipsum}		
\usepackage{graphicx}
\usepackage{caption}
\usepackage{subcaption}
\usepackage{natbib}
\usepackage{doi}
\usepackage{xcolor}
\usepackage{multirow}
\usepackage{blindtext}
\usepackage{titlesec}
\usepackage{algorithm} 
\usepackage{algpseudocode}
\usepackage{tabularx}
\usepackage[acronym]{glossaries}
\glsdisablehyper
\newacronym{ame}{AME}{attention and memory enhancement}
\newacronym{bert}{BERT}{bidirectional encoder representations from transformers}
\newacronym{boot}{BooT}{bootstrapped transformer}
\newacronym{cnn}{CNN}{convolutional neural network}
\newacronym{cv}{CV}{computer vision}
\newacronym{dag}{DAG}{directed acyclic graph}
\newacronym{dgerd}{DGERD}{disjunctive graph embedded recurrent decoding}
\newacronym{dt}{DT}{decision transformer}
\newacronym{esper}{ESPER}{environment-stochasticity-independent representations}
\newacronym{gan}{GAN}{generative adversarial network}
\newacronym{gnn}{GNN}{graph neural network}
\newacronym{gpt}{GPT}{generative pre-trained transformer}
\newacronym{gru}{GRU}{gated recurrent unit}
\newacronym{gtr-xl}{GTrXL}{gated transformer-XL}
\newacronym{happo}{HAPPO}{heterogeneous-agent proximal policy optimization}
\newacronym{hpo}{HPO}{hyper-parameter optimization}
\newacronym{iid}{i.i.d}{independent and identically distributed}
\newacronym{iot}{IoT}{internet of things}
\newacronym{iris}{IRIS}{imagination with auto-regression over an inner speech}
\newacronym{llm}{LLM}{large language model}
\newacronym{lstm}{LSTM}{long short-term memory}
\newacronym{maans}{MAANS}{multi-agent active neural SLAM}
\newacronym{maml}{MAML}{model-agnostic meta-learning}
\newacronym{mappo}{MAPPO}{multi-agent proximal policy optimization}
\newacronym{marl}{MARL}{multi-agent reinforcement learning}
\newacronym{mbrl}{MBRL}{model-based reinforcement learning}
\newacronym[longplural=Markov decision processes]{mdp}{MDP}{Markov decision process}
\newacronym{mlp}{MLP}{multi-layer perceptron}
\newacronym{mse}{MSE}{mean-squared error}
\newacronym{mtrl}{MTRL}{multi-task reinforcement learning}
\newacronym{nlp}{NLP}{natural language processing}
\newacronym{pomdp}{POMDP}{partially observable Markov decision process}
\newacronym{ppo}{PPO}{proximal policy optimization}
\newacronym{qdt}{QDT}{$Q$-learning decision transformer}
\newacronym{rat}{RAT}{relation-aware transformer}
\newacronym{rl}{RL}{reinforcement learning}
\newacronym{rlhf}{RLHF}{reinforcement learning with human feedback}
\newacronym{rnn}{RNN}{recurrent neural network}
\newacronym{td}{TD}{temporal difference}
\newacronym{tr-xl}{TrXL}{transformer-XL}
\newacronym{tt}{TT}{trajectory transformer}
\newacronym{vit}{ViT}{vision transformer}
\newacronym{dtd}{DTd}{decision transducer}

\graphicspath{ {./figures/} }

\begin{document}

\title{Transformers in Reinforcement Learning: A Survey}

\author{Pranav Agarwal}
\affiliation{%
 \institution{École de Technologie Supérieure/Mila}
 \city{Montréal}
 \state{Québec}
 \country{Canada}}
\email{pranav.agarwal.1@ens.etsmtl.net}

\author{Aamer Abdul Rahman}
\affiliation{%
 \institution{École de Technologie Supérieure/Mila}
 \city{Montréal}
 \state{Québec}
 \country{Canada}}
\email{aamer.abdul-rahman.1@ens.etsmtl.net}

\author{Pierre-Luc St-Charles}
\affiliation{
 \institution{Mila, Applied ML Research Team}
 \city{Montréal}
 \state{Québec}
 \country{Canada}}
\email{pierreluc.stcharles@mila.quebec}

\author{Simon J.D. Prince}
\affiliation{%
 \institution{University of Bath}
 \city{Bath}
 \country{United Kingdom}}
\email{sjdp23@bath.ac.uk}

\author{Samira Ebrahimi Kahou}
\affiliation{%
 \institution{École de Technologie Supérieure/Mila/CIFAR}
 \city{Montréal}
 \state{Québec}
 \country{Canada}}
\email{samira.ebrahimi.kahou@gmail.com} 

\renewcommand{\shortauthors}{Agarwal, Rahman, St-Charles, Prince, and Ebrahimi Kahou}

\begin{abstract}

Transformers have significantly impacted domains like natural language processing, computer vision, and robotics, where they improve performance compared to other neural networks. This survey explores how transformers are used in \gls{rl}, where they are seen as a promising solution for addressing challenges such as unstable training, credit assignment, lack of interpretability, and partial observability. We begin by providing a brief domain overview of \gls{rl}, followed by a discussion on the challenges of classical \gls{rl} algorithms. Next, we delve into the properties of the transformer and its variants and discuss the characteristics that make them well-suited to address the challenges inherent in \gls{rl}. We examine the application of transformers to various aspects of \gls{rl}, including representation learning, transition and reward function modeling, and policy optimization. We also discuss recent research that aims to enhance the interpretability and efficiency of transformers in \gls{rl}, using visualization techniques and efficient training strategies. Often, the transformer architecture must be tailored to the specific needs of a given application. We present a broad overview of how transformers have been adapted for several applications, including robotics, medicine, language modeling, cloud computing, and combinatorial optimization. We conclude by discussing the limitations of using transformers in \gls{rl} and assess their potential for catalyzing future breakthroughs in this field.
\end{abstract}



\maketitle

\newpage
\section{Introduction}
\label{sec:introduction}

\Glsfirst{rl} is a learning paradigm that enables sequential decision-making by learning from feedback obtained through trial and error. It is usually formalized in terms of an \gls{mdp}, which provides a mathematical framework for modeling the interaction between an agent and its environment. 

Most \gls{rl} algorithms optimize the agent's policy to select actions that maximize the expected cumulative reward. In deep \gls{rl}, neural networks are used as function approximators for mapping the current state of the environment to the next action and for estimating future returns. This approach is beneficial when dealing with large or continuous state spaces that make tabular methods computationally expensive~\citep{sutton2018reinforcement} and has been successful in challenging applications~\citep{arulkumaran2017deep, nguyen2020deep, latif2022survey}. However, standard neural network architectures like \glspl{cnn} and \glspl{rnn} struggle with long-standing problems in \gls{rl}. These problems include partial observability~\citep{esslinger2022deep}, inability to handle high-dimensional state and action spaces~\citep{barto2003recent}, and difficulty in handling long-term dependencies~\citep{Chen2022TransDreamerRL}.

Partial observability is a challenge in \gls{rl}~\citep{liu2022partially}; in the absence of complete information, the agent may be unable to make optimal decisions. A typical way to address this problem is to integrate the agent's input~\citep{shao2019survey} over time using \glspl{cnn} and \glspl{rnn}. However, \glspl{rnn} tend to forget information~\citep{pascanu2013difficulty}, while \glspl{cnn} are limited in the number of past-time steps they can process~\citep{karpathy2014large}. Various strategies have been proposed to overcome this limitation, including gating mechanisms, gradient clipping, non-saturating activation functions, and manipulating gradient propagation paths~\citep{Ribeiro2019BeyondEA}.
Sometimes different data modalities, such as text, audio, and images are combined to provide additional information to the agent~\citep{lathuiliere2019neural, song2021multimodal, carta2020visualhints}. However, integrating encoders for different modalities increases the model's architectural complexity. With \glspl{cnn} and \glspl{rnn}, it is also difficult to determine which past actions contributed to current rewards~\citep{ma2021longterm}. This is known as the credit assignment problem. These challenges and others, such as training instability, limit the scope of most \gls{rl} applications to unrealistic virtual environments.

The transformer was first introduced in 2017~\citep{vaswani2017attention} and has rapidly impacted the field of deep learning~\citep{lin2022survey}, improving the state-of-the-art in \gls{nlp} and \gls{cv} tasks~\citep{tunstall2022natural, khan2022Transformers, devlin2018bert, petit2021u, zhong2020self}. The key idea behind this neural network architecture is to use a self-attention mechanism to capture long-range relationships within the data. This ability to model large-scale context across sequences initially made transformers well-suited for machine translation tasks. Transformers have since been adapted to tackle more complex tasks like image segmentation~\citep{petit2021u}, visual question answering~\citep{zhong2020self}, and speech recognition~\citep{dong2018speech}. 

\begin{figure}[ht]
\includegraphics[scale=0.9]{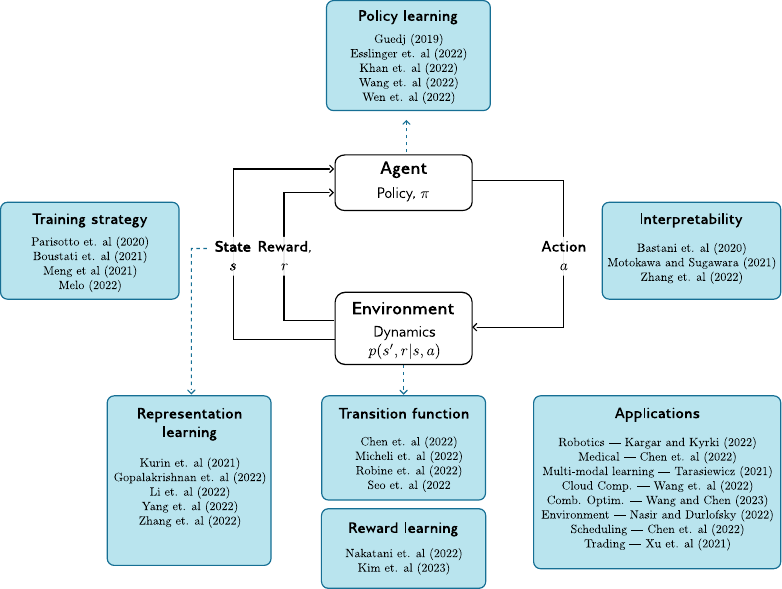}
\centering
\caption{This survey presents a comprehensive overview of the use of transformers in \gls{rl}. Modeling an \gls{rl} policy may involve representation learning, modeling the transition function, reward function learning, and policy learning. Transformers can be used across all of these tasks.}
\label{flow_chart}
\end{figure}

This document surveys the use of transformers in \gls{rl}. We begin by providing a concise overview of \gls{rl} (Sec.~\ref{sec:background:rl}) and transformers (Sec.~\ref{sec:background:tran}) that is accessible to readers with a general background in machine learning. We highlight challenges that classical \gls{rl} approaches face and how transformers can help deal with these challenges (Sec.~\ref{sec:background:chall} and~\ref{sec:advant:tran}). Transformers can be applied to \gls{rl} in different ways (Fig.~\ref{flow_chart}). We discuss how they can be used to learn representations (Sec.~\ref{sec3:representation}), model transition functions (Sec.~\ref{sec3:world}), learn reward functions (Sec.~\ref{sec3:reward}) and learn policies (Sec.~\ref{sec3:policy}). In Sec. ~\ref{sec:trainstra} and Sec. ~\ref{sec:explainab}, we discuss different training and interpretation strategies, and in Sec.~\ref{sec:applic}, we provide an overview of \gls{rl} applications that use transformers, including robotics, medicine, language modeling, edge-cloud computing, combinatorial optimization, environmental sciences, scheduling, trading, and hyper-parameter optimization. Finally, we discuss limitations and open questions for future research (Sec.~\ref{sec:limit}). With this work, we aim to inspire further research and facilitate the development of \gls{rl} approaches for real-world applications.

\section{Background}
\label{sec:background}
This section introduces the fundamental concepts of \gls{rl} and discusses associated challenges. We also provide an overview of transformers and their potential advantages in \gls{rl}. 

\subsection{Reinforcement Learning}
\label{sec:background:rl}

\Glsfirst{rl} is a reward-based learning paradigm that enables agents to learn from their experience and improve their performance over time. This is commonly formulated in terms of \acrfullpl{mdp}, in which the agent chooses an action~$(\mathbf{a} \in \mathbf{A})$ based on the state~$(\mathbf{s} \in \mathbf{S})$ of the environment and receives feedback in the form of rewards~$(r \in \mathbb{R})$. The \gls{mdp} framework assumes that the environment satisfies the Markov property, which asserts that the next state is independent of the past states, given the present state and the most recent action. This allows the agent to make decisions based only on the current state without tracking the history of previous states and actions. After taking action, the agent receives feedback from the environment in the form of a reward. The state is updated to a new value~$(\mathbf{s}^\prime \in \mathbf{S})$, as determined by the transition function~$p(\mathbf{s'}\mid \mathbf{s},\mathbf{a})$ describing how the environment responds to the agent's action. The overall goal of \gls{rl} is to learn how to solve a multi-step problem by maximizing the return~$g$, which is the total discounted reward:
\begin{equation}
\label{eq:1}
 g = \sum_{t=0}^{T} \gamma^t \cdot r_t,
\end{equation}
\noindent where~$r_t$ is the reward received at each time step~$t$,~$T$ is the total number of time steps in the given episode, and~$\gamma\in[0,1]$ is a discount factor, affecting the importance given to immediate rewards versus future rewards in a given episode.

There are multiple categories of \gls{rl} algorithms, each with advantages and disadvantages~\citep{almahamid2021reinforcement}. Choosing the appropriate category relies on factors such as the problem's complexity, the size of the state and action spaces, and available computational resources. We now briefly review these categories. \newline

\noindent \textbf{Model-Based \gls{rl}}. In model-based \gls{rl}, a transition function is learned using transitions ($\mathbf{s}, \mathbf{a}, r, \mathbf{s}'$) generated by environmental interaction. This transition function models the probability distribution~$p(\mathbf{s}', r \mid \mathbf{s}, \mathbf{a})$ over the subsequent state~$\mathbf{s}'$ and reward~$r$ given the current state~$\mathbf{s}$ and action~$\mathbf{a}$. By leveraging this learned model, the agent plans and selects actions that maximize the expected return~$g$. However, this approach can be computationally expensive and may suffer from inaccuracies in the learned model, leading to sub-optimal performance~\citep{moerland2020model}. Despite these drawbacks, this approach is sample efficient. In other words, it can achieve good performance using relatively few interactions with the environment compared to other methods~\citep{mohanty2021measuring, wang2022sample}.
\newline

\noindent \textbf{Model-Free \gls{rl}}. In model-free \gls{rl}, optimal actions are learned by direct interaction with the environment. Methods from this category cannot model state-transition dynamics to plan actions, which can result in slower convergence and lower sample efficiency compared to model-based \gls{rl}~\citep{yarats2021improving}.
However, model-free \gls{rl} is more adaptable to environmental changes, making it more robust in complex or noisy environments~\citep{mnih2013playing, lillicrap2015continuous, schulman2017proximal}. Additionally, it is less computationally expensive as it does not need to learn a model of the environment. 

Further, \gls{rl} methods can be categorized as on-policy, off-policy, and offline \gls{rl} depending on how the data collection relates to the policy being learned.
\newline

\noindent \textbf{On-Policy \gls{rl}}. On-policy \gls{rl} approaches use the current policy to gather transitions for updating the value function. For instance in SARSA~\citep{sutton2018reinforcement}, the current policy is used to collect a tuple~$(\mathbf{s, a}, r, \mathbf{s', a'})$, consisting of the current state-action pair~$\mathbf{s, a}$, the immediate reward~$r$, and the next state-action pair~$\mathbf{s', a'}$. This is then used for estimating the return of the current state-action pair~$Q_{\text{target}}(\mathbf{s, a})$ and the next state-action pair~$Q_{\text{target}}(\mathbf{s', a'})$. The value function~$Q$ is updated using the following \glsfirst{td} learning rule:
\begin{equation}
 Q_{\text{target}}(\mathbf{s},\mathbf{a}) \leftarrow Q_{\text{target}}(\mathbf{s},\mathbf{a}) + \alpha \cdot \left[ r + \gamma \cdot Q_{\text{target}}(\mathbf{s}',\mathbf{a}') - Q_{\text{target}}(\mathbf{s},\mathbf{a}) \right],
\end{equation}
where~$\alpha\in(0,1]$ represents the learning rate. 
Although on-policy methods are comparatively easy to implement, they have several drawbacks. They tend to be sample inefficient~\citep{larsen2021comparing}, requiring significant interaction with the environment to achieve good performance. Additionally, they can be susceptible to policy oscillation and instability~\citep{young2020understanding}, and they lack the flexibility to explore, slowing down the learning process and resulting in sub-optimal policies. \newline

\noindent \textbf{Off-Policy \gls{rl}}. Off-policy \gls{rl} strategies use two policies --- a behavior policy and a target policy. The behavior policy collects data that is subsequently used to estimate the expected return of an action under the given target policy. Since the behavior policy is used for data collection, it can explore different states and actions without affecting the current target policy. Thus, off-policy methods are well-suited for understanding the value of a given state and action. Usually, the target policy is updated using the behavior policy via \textit{importance sampling (IS)}. This adjusts the value estimates of the target policy based on the IS ratio between the behavior~$b(\mathbf{a} | \mathbf{s})$ and target~$\pi(\mathbf{a} | \mathbf{s})$ policies:

\begin{equation}
 \text{IS} = \frac{\pi(\mathbf{a}|\mathbf{s})}{b(\mathbf{a}|\mathbf{s})},
\end{equation}
\noindent and the final value estimate is given as:
\begin{equation}
 Q_{\text{target}}(\mathbf{s},\mathbf{a}) \leftarrow Q_{\text{target}}(\mathbf{s},\mathbf{a}) + \alpha \cdot \left[ \text{IS} \cdot(r + \gamma \cdot Q_{\text{target}}(\mathbf{s}',\mathbf{a}')) - Q_{\text{target}}(\mathbf{s},\mathbf{a}) \right].\newline 
\end{equation}

\noindent \textbf{Offline-RL.} Offline \gls{rl} or batch \gls{rl} \cite{Levine2020OfflineRL} uses a static dataset of transitions, denoted by~$\mathbf{D}=\{\mathbf{s_t, a_t}, r_t, \mathbf{s_{t}^{'}}\}_{t=1}^T$, collected using a behavior policy, and so does not require interaction with the environment to collect trajectories. Offline-RL updates the state-action value function~$Q_{\text{target}}$ as:

\begin{equation}
 Q_{\text{target}}(\mathbf{s,a}) \leftarrow Q_{\text{target}}(\mathbf{s,a}) + \alpha \cdot \left[ r + \gamma \cdot \max_{\mathbf{a'}} (Q_{\text{target}}(\mathbf{s',a'})) - Q_{\text{target}}(\mathbf{s,a}) \right],
\end{equation}

\noindent where the max operator estimates the maximum expected return over all possible actions in the next state. Offline \gls{rl} is a more practical strategy for safety-critical applications as it does not require interactions with the environment~\citep{shi2021offline, killian2023risk}. However, the static nature of the dataset does not allow the agent to explore and adapt to new information, potentially limiting performance~\citep{lu2022challenges}. \newline

\noindent \textbf{Multi-Agent Reinforcement Learning.} Online, offline, and off-policy learning setups can be used to facilitate adaptive decision-making in dynamic environments with multiple interacting agents~\citep{gronauer2022multi}. Each of the~$I$ agents has its own policy~$\pi_i(\mathbf{a}_i|\mathbf{s}_i)$, state space~$\mathbf{S}_i$, and action space~$\mathbf{A}_i$. The agents interact with each other and the environment, and their actions can affect the outcomes of other agents. The goal in \gls{marl} is to learn a joint policy~$\pi(\mathbf{a}_1, \mathbf{a}_2, ..., \mathbf{a}_I|\mathbf{s}_1, \mathbf{s}_2, ..., \mathbf{s}_I)$ that maximizes the collective reward of all agents. Formally, the objective in \gls{marl} can be expressed as maximizing the expected sum of discounted rewards of all the agents:
\begin{equation}
\mathbb{E}\left[\sum_{t=0}^{\infty}\gamma^t\sum_{i=1}^I r_i^{(t)}\right],
\end{equation}
\noindent where~$\gamma$ is the discount factor and~$r_i^{(t)}$ is the reward received by agent~$i$ at time~$t$.
In general, there are two main approaches to \gls{marl}: decentralized policies~\citep{pmlr-v80-zhang18n}, and centralized training~\citep{sharma2021survey}. Decentralized policies involve each agent independently learning its policy without explicit coordination or communication with other agents. In contrast, centralized training uses a shared value function that considers joint states and actions, enabling agents to communicate and coordinate their actions through a communication protocol. Communication protocols~\citep{foerster2016learning} facilitate information exchange and collaboration among agents. \newline

\noindent \textbf{Upside Down RL.} Classical \gls{rl} usually involves optimizing policies by estimating the expected future return. Upside-down \gls{rl}~\citep{schmidhuber2019reinforcement, arulkumaran2022all} flips the traditional \gls{rl} paradigm and uses the desired return~$g$, the horizon~$h$ (i.e., the time remaining until the end of the current trial), and the state as inputs. This input acts as a {\em command} which is mapped to action probabilities. Upside-down \gls{rl} offers improved stability compared to classical \gls{rl} as it avoids the need to estimate the value function, which can introduce instabilities in traditional \gls{rl} algorithms~\citep{chen2021decisiontransformer, sutton2018reinforcement}. The loss function of upside-down \gls{rl} can be defined as:
\begin{equation}
 \mathcal{L}(\boldsymbol\theta) = \sum_{t=0}^{} \Bigl[ \mathbf{a}_t - f(\mathbf{s}_t, g_t, h_t,\boldsymbol\theta) \Bigr]^2,
\end{equation}
where~$\boldsymbol\theta$ contains the model parameters. The term~$\mathbf{a}_t$ is the action at time step~$t$, and~$f(\mathbf{s}_t, g_t, h_t,\boldsymbol\theta)$ is the predicted action when conditioned on state~$\mathbf{s}_t$, expected future return~$g_t$, and horizon~$h_t$. 

\subsection{Challenges in Reinforcement Learning}
\label{sec:background:chall}
In this section, we discuss the different challenges of classical \gls{rl} algorithms. \newline

\noindent{\textbf{Curse of Dimensionality}}. Real-world applications often involve high-dimensional state spaces, which makes it hard for classical \gls{rl} algorithms to learn optimal policies~\citep{barto2003recent}. This is because the required training data grows exponentially as the data dimensionality increases~\citep{wang2020breaking}. One way to mitigate this problem is to encode high-dimensional states into a lower-dimensional space. \gls{rl} policies perform better when trained on encoded low-dimensional data~\citep{yarats2021improving}. \newline

\noindent{\textbf{Partially Observable Environment}}. A partially observable environment presents a challenge for \gls{rl} algorithms, as the agent cannot access observations that contain complete information about the environmental state at each time-step~\citep{vinyals2019grandmaster, akkaya2019solving}. Without complete information, the algorithm may struggle to make the best decision, leading to uncertainty and compromises in performance. To address this, the policy must maintain an internal representation of the state, often in the form of memory, from which the actual state can be estimated~\citep{icarte2020act}. Historically, this has often been done with \glspl{rnn}, but these cannot efficiently model long contexts~\citep{pascanu2013difficulty, hochreiter1998vanishing}. \newline

\noindent{\textbf{Credit Assignment}}. The term credit assignment refers to the problem of associating the actions taken by an agent with the reward it receives~\citep{mesnard2020counterfactual}. This is challenging for two reasons: First, the reward may be delayed; the agent may not be able to observe the consequences of its actions until several time steps into the future. Second, other factors or multiple actions may influence the received reward, making it challenging to identify which action led to that reward.
Inaccurate credit assignment can lead to slower training and sub-optimal policies~\citep{ausin2021tackling}. Moreover, when the reward is sparse (i.e., when the agent receives little feedback for its actions), the credit-assignment problem becomes even more difficult~\citep{seo2019rewards}. One potential solution is to use models that integrate information across all time steps, which may be better suited for solving this issue~\citep{chen2021decisiontransformer}. \newline

\noindent Recent studies have exploited transformers to tackle these three key challenges. Transformers have demonstrated success in modeling long-term dependencies in sequential data while showing promising results in promoting generalization and faster learning in domains such as \gls{nlp} and \gls{cv}. We now provide a brief overview of transformers and explore the various ways in which they have been applied to learning optimal \gls{rl} policies. 

\subsection{Transformers}
\label{sec:background:tran}

\begin{figure}[ht]
\includegraphics[width=0.75\linewidth]{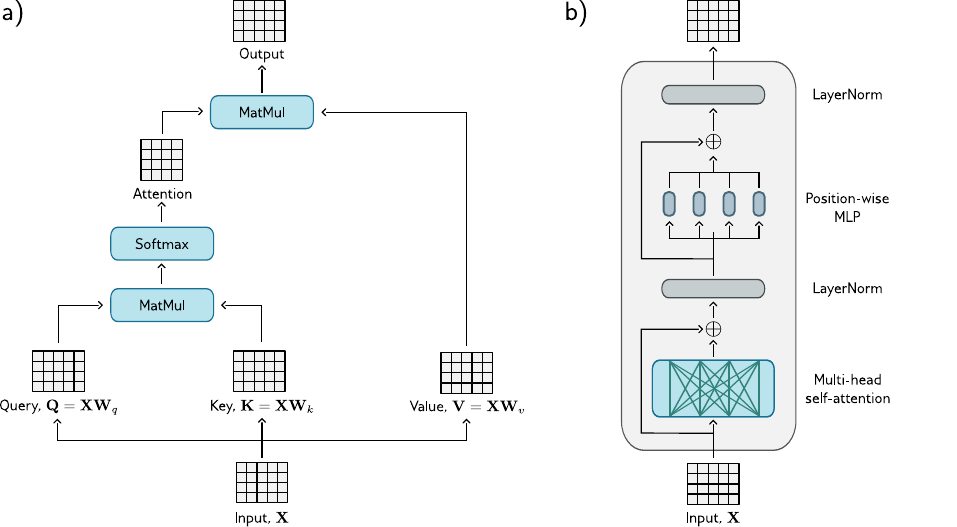}
\centering
\caption{a) The dot-product attention mechanism (for 4 embeddings of size 5 each). Input features ($\mathbf{X}$) are processed using the key ($\mathbf{K}$), query ($\mathbf{Q}$), and value ($\mathbf{V}$) tensors. Each query undergoes a dot product with every key, and the result is normalized using the softmax operator to compute the attention. This is then used to weigh the values to produce the output. b) The (multi-head) self-attention mechanism is just one component in the transformer block, which also contains residual links, layer normalization, and parallel \glspl{mlp} that process each input embedding separately.
}
\label{dot-product}
\end{figure}

Transformers are a class of neural network architectures consisting of multiple layers, each containing a multi-head self-attention mechanism, parallel fully-connected networks, residual connections, and layer normalization. Given a sequence of~$N$ input embeddings, transformers produce a sequence of~$N$ output embeddings, each of which represents the relationship between the corresponding input embedding and the rest of the input sequence. In \gls{nlp}, the input embeddings may represent words from a given sentence, while in \gls{rl}, they may represent different states.

The {\em self-attention mechanism} allows each input embedding (each row of~$\mathbf{X}$) to simultaneously attend to all the other embeddings in the input sequence. It computes one attention score for each pair of inputs. This is done by projecting each input into a query~$\mathbf{Q}=\mathbf{XW}_q$ and a key~$\mathbf{K}=\mathbf{XW}_k$ tensor. The attention scores are then computed by taking the dot product of each query vector (row of the query tensor) with every key vector (row of the key tensor), followed by a softmax operation that normalizes the resulting scores such that they add up to one for each query. The attention scores are then used to compute a weighted sum of value~$\mathbf{V}=\mathbf{XW}_v$ tensors (see Fig.~\ref{dot-product}): 
\begin{equation}
\text{Attention}(\mathbf{Q, K, V}) =\operatorname{Softmax}\left( \frac{\mathbf{Q} \mathbf{K}^{\top}}{\sqrt{d_q}}\right) \cdot \mathbf{V}.
\end{equation}

\noindent To help stabilize gradients during training, the dot product is scaled by a factor of~$\sqrt{d_{q}}$, where~$d_q$ is the dimension of the query tensor. 

Transformers often compute multiple sets of attention scores in parallel (each with a different set of learned parameters~$\mathbf{W}_{k}, \mathbf{W}_{q}, \mathbf{W}_{v}$). This allows the model to attend to multiple aspects of the input sequence simultaneously and is known as \emph{multi-head self-attention}. Each attention ``head'' output is concatenated and linearly transformed to produce the final output representation. For applications where the order of the inputs is important, a position encoding is added that allows the network to establish the position of each input.

\begin{figure}[ht]
\includegraphics[width=1.0\linewidth]{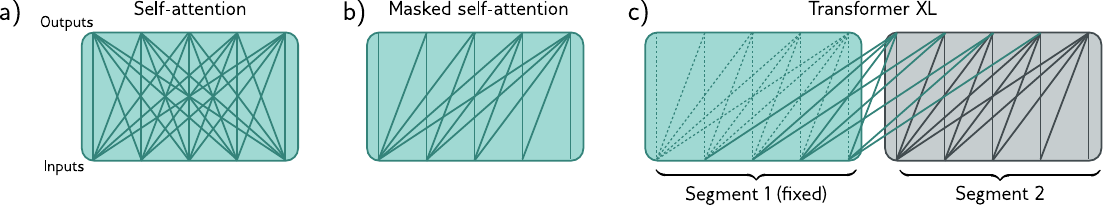}
\centering
\caption{a) {\em Self-attention} computes attention scores for a given segment by considering all input across all time steps. b) {\em Masked self-attention} considers the current and past time steps while disregarding the future. c) {\em Transformer XL}~\citep{dai2019Transformer} establishes a connection between past segments, incorporating information from multiple previous segments. It calculates attention scores by combining the hidden state of the previous segment, enabling the modeling of longer-term dependencies.}
\label{tr-xl}
\end{figure}

In a transformer block (Fig.~\ref{dot-product}b), a residual connection is placed around the multi-head self-attention mechanism. This improves training stability by allowing the gradient to flow easily through the network. The output is then processed using layer normalization, which normalizes the activations of each layer across the feature dimension. Each output is processed in parallel by the same \gls{mlp}. Once more, these are bypassed by a residual connection, and a second layer norm is subsequently added. \newline

\begin{figure}[ht]
\includegraphics[width=0.8\linewidth]{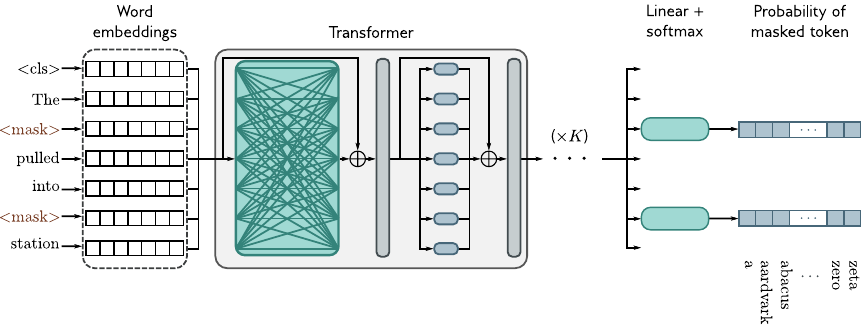}
\centering
\caption{The \gls{bert} architecture. The input tokens are converted to word embeddings and passed through transformer layers to produce latent representations. Some input tokens are replaced with mask embedding as part of the formulation for the pre-training task. The output embeddings predict the missing (masked) word using a softmax and multi-class classification loss. In each loss term corresponding to the prediction of a masked word, \gls{bert} uses context from before and after the word.
}
\label{bert}
\end{figure}

\noindent{\textbf{Architectural Variations.}} The \gls{bert} model~\citep{devlin2018bert} and the \gls{gpt}~\citep{radford2018improving} are two popular variants of the transformer architecture. \Gls{bert} (Fig.~\ref{bert}) is a {\em transformer encoder} in which each output receives information from every input in the self-attention mechanism (Fig.~\ref{tr-xl}a). The goal is to process the incoming data to generate a latent representation that integrates contextual information. This can be particularly useful in \gls{rl}, as it enables the agent to make informed decisions based on a more comprehensive understanding of the environment~\citep{banino2021coberl, wang2023solving}. 

\begin{figure}[ht]
\includegraphics[width=0.8\linewidth]{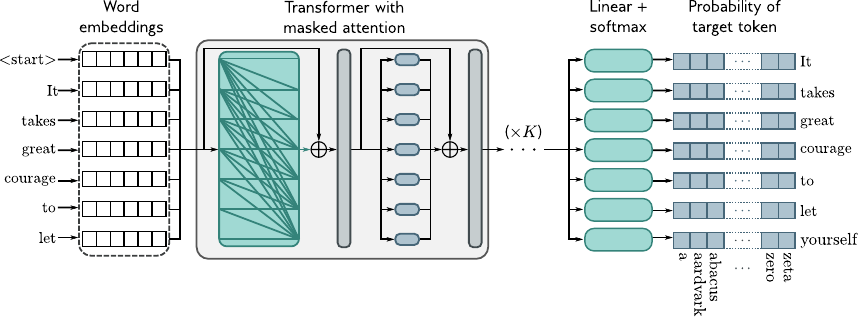}
\centering
\caption{The \gls{gpt} architecture. The tokens are mapped to embeddings through transformer layers with masked self-attention, such that each position attends only to earlier tokens. The goal is to predict the next token at each position correctly. This approach is efficient to train as every word contributes to the final loss.}
\label{gpt}
\end{figure}

Conversely, \gls{gpt} (Fig.~\ref{gpt}) uses a {\em decoder architecture} to auto-regressively generate a sequence of output tokens, considering only the past tokens. The use of masked self-attention (Fig.~\ref{tr-xl}b) prevents it from cheating by looking ahead to tokens that it should not know yet during training by clamping the associated attention values to zero. In \gls{rl}, this autoregressive nature can be used to implement an \gls{rl} policy that is conditioned on a sequence of past states and actions~\citep{olmo2021gpt3, chen2021decisiontransformer, Janner2021ReinforcementLA}. \Gls{gpt} uses multiple blocks, each containing a multi-head attention mechanism. The original transformer~\citep{vaswani2017attention} combined these two approaches in an encoder-decoder architecture for machine translation. An encoder architecture processes the incoming sentence, and the decoder auto-regressively produces the output sentence. In doing so, the decoder also considers the attention over the encoder's latent representation using a ``cross-attention'' block.\newline

\noindent{\textbf{Vision Transformers.}} Inspired by the success of transformer-based architectures like \gls{bert} and \gls{gpt} in \gls{nlp}, \citet{dosovitskiy2020image} proposed the \gls{vit} architecture for processing images. \Gls{vit} architecture~\citep{dosovitskiy2020image} is suited to a wide range of \gls{rl} tasks where images must be used to learn a policy~\citep{tao2022evaluating, kargar2021vision, goulao2022pretraining}. The \gls{vit} architecture is a transformer encoder that processes patches of the image (Fig.~\ref{fig:representation}). Each patch is combined with a positional encoding that provides knowledge about its original image location.\newline 

\noindent{\textbf{Transformer-XL.}} The computational complexity of the self-attention mechanism increases quadratically with input sequence length due to an exponential increase in pairwise comparison. Hence, the transformer architectures discussed so far typically partition long input sequences into shorter sequences to reduce memory demands. While this approach helps minimize memory usage, it makes capturing global context challenging. Moreover, the conventional transformer model is limited because it does not consider the boundaries of the input sequence when forming context. Instead, it selects consecutive chunks of symbols without regard for a sentence or semantic boundaries. This can result in context fragmentation, where the model lacks the necessary contextual information to accurately predict the first few symbols in a sequence. The \gls{tr-xl} architecture~\citep{dai2019Transformer} addresses these issues by dividing the input into segments and incorporating segment-level recurrence (Fig.~\ref{tr-xl}c) and relative positional encodings. By caching and reusing the representation computed for the previous segment during training, \gls{tr-xl} can extend the context and better capture long-term dependencies. Additionally, \gls{tr-xl} can process elements of new segments without recomputing the past segments, leading to faster inference.

\subsection{Key Advantages of Transformers in RL}
\label{sec:advant:tran}
This section outlines the transformer characteristics that are important for \gls{rl} applications. \newline

\noindent{\textbf{Attention Mechanism.}} The attention mechanism is crucial in transformers for sequence modeling of states~\citep{benjamins2022contextualize}. It enables the \gls{rl} agent to focus selectively on relevant cues in the environment~\citep{manchin2019reinforcement} and ignore redundant features, which leads to faster training. This is particularly useful in high-dimensional state spaces where there are a large number of input elements. \newline

\noindent{\textbf{Multi-Modal Architectures.}} For complex tasks, \gls{rl} agents may require additional information from different data modalities~\citep{zhang2018multimodal, kiran2021deep}. Past approaches have used different architectures to handle multiple modalities~\citep{ramachandram2017deep}. However, transformers can process multiple modalities of data (e.g., text, images) effectively~\citep{jaegle2021perceivergeneral, xu2022multimodal} using the same architecture. \newline

\noindent{\textbf{Parallel Processing.}} Learning a policy in \gls{rl} can be computationally expensive, especially for complex tasks requiring many samples~\citep{ceron2021revisiting}. \Glspl{rnn} require the sequential processing of inputs, which is inefficient. Transformers are well-suited for parallelization due to their self-attention mechanism, which considers all inputs simultaneously. \Gls{rl} algorithms can exploit transformers to learn more efficient policies in significantly less time. \newline

\noindent{\textbf{Scalability.}} Current \gls{rl} algorithms struggle to scale effectively to complex tasks that require the integration of multiple skills~\citep{zhan2017scalable, kalashnikov2018scalable}. However, the performance of transformers has been shown to improve smoothly as the size of the model, dataset, and compute increases~\citep{kaplan2020scaling}. This ability can potentially be leveraged in \gls{rl} to create generalist agents capable of performing various tasks in different environments and with different embodiments~\citep{LeeMultiGameDT}. \newline

These points highlight how the properties of transformers make them attractive for \gls{rl}. In the following sections, we examine the use of transformers in each stage of the \gls{rl} workflow, including representation learning (Sec.~\ref{sec3:representation}), policy learning (Sec.~\ref{sec3:policy}), learning the reward function (Sec.~\ref{sec3:reward}), and modeling the environment (Sec.~\ref{sec3:world}). Additionally, we cover various training strategies (Sec.~\ref{sec:trainstra}) and techniques for interpreting \gls{rl} policies that use transformers (Sec.~\ref{sec:explainab}).

\section{Representation Learning}
\label{sec3:representation}
Concise and meaningful representations are critical for efficient decision-making~\citep{lesort2018state} in \gls{rl}. Empirical evidence has shown that training agents directly on high dimensional data, such as image pixels, is sample inefficient~\citep{Lake2016BuildingMT}. Therefore, good data representations are crucial for learning \gls{rl} policies~\citep{lesort2018state, laskin2020curl} since they can enhance performance, convergence speed, and policy stability~\citep{ghosh2020representations}.

\begin{figure}[h]
\includegraphics[scale=0.90]{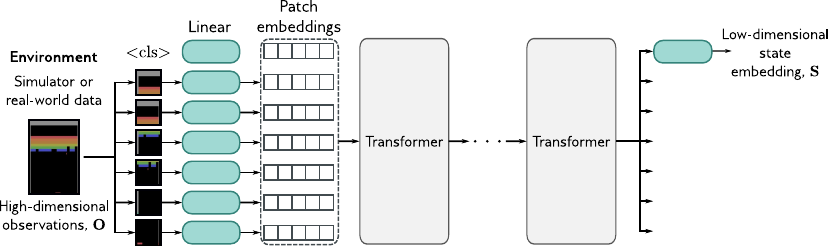}
\centering
\caption{Representation learning simplifies the information given to \gls{rl} agents by condensing the original high-dimensional input (here a frame from an Atari game). Transformers can learn low-dimensional representations~$\mathbf{s}$ from high-dimensional observations~$\mathbf{o}$. In this case, this is achieved by employing a vision transformer. The input is provided in multiple patches, and the transformer encoder learns a shared low-dimensional representation contextualized using a \textit{<cls>} input token.}
\label{fig:representation}
\end{figure}

For instance, in self-driving cars, the raw sensory inputs~$\mathbf{o}_t$ (e.g., camera images, LIDAR readings) are high dimensional and often contain redundant information. If these inputs~$\mathbf{o}_t$ are mapped to a compact representation~$\mathbf{s}_t$, the \gls{rl} agent can learn more efficiently. Similarly, in game-playing scenarios (Fig.~\ref{fig:representation}), it's helpful to encode and extract relevant features from the pixel to be used as input to the \gls{rl} algorithm learning the policy. Transformers can produce transferable and discriminative feature representations for diverse data modalities~\citep{Zhou2021ConvNetsVT, Brown2020LanguageMA, ZhangVisRep, Choi2020EncodingMS, Ying2021DoTR}. 

\subsection{Comparisons between Transformers, CNNs, and GNNs} 

Encoding high-dimensional representations using pre-trained \gls{cnn} and transformer architectures is an active research area. Both approaches yield comparable performance in computer vision tasks~\citep{woo2023convnext, liu2022convnet, malpure2021investigating}. However, several studies~\citep{Zhou2021ConvNetsVT, ZhangVisRep} have shown that transformers generate more expressive representations than \glspl{cnn} for tasks where the data distribution differs at training and test time. This advantage stems from \gls{cnn}'s inherent inductive bias towards local spatial features, which limits their ability to capture the global dependencies necessary for reasoning~\citep{vo2017multi}. Transformers can encode the image as a sequence of patches without local convolution and resolution reduction (Fig.~\ref{fig:representation}). Hence, they model the global context in every layer, leading to a stronger representation for learning efficient policies~\citep{dosovitskiy2020image}. Transformers exhibit comparable generalization capabilities to \glspl{gnn} in graphs~\citep{dwivedi2020generalization} and, in some instances, outperform them by capturing long-range semantics~\citep{Ying2021DoTR}. 

\Gls{mtrl} is a learning paradigm where an agent is trained to perform multiple tasks simultaneously. It has traditionally relied on \glspl{gnn} to handle incompatible environments (i.e., differing state-action spaces)~\citep{Wang2018NerveNetLS, Huang2020OnePT}. This is due to the ability of \glspl{gnn} to operate on graphs of variable sizes. However, \citet{Kurin2021MyBI} hypothesize that the restrictive nature of message passing in sparse graphs may adversely impact performance. They propose replacing \glspl{gnn} with transformers which obviates the need to learn multi-hop communication; the transformer can be considered a \gls{gnn} applied to fully-connected graphs with attention as an edge-to-vertex aggregation operation~\citep{battaglia2018relational}. This enables a dedicated message-passing scheme for each state and pass, effectively avoiding the requirement for multi-hop message propagation. This overcomes the challenges of gradient propagation and information loss arising from such multi-hop propagation. The transformer-based model of~\citet{Kurin2021MyBI}, Amorpheus, learns better representations and improves performance without imposing a relational inductive bias. 

\subsection{Advanced Representation Learning using Transformers}
Transformers, in combination with other attention mechanisms, enable the learning of expressive representations. SloTTAr~\citep{Gopalakrishnan2022UnsupervisedLO} combines a transformer encoder-decoder architecture with slot attention~\citep{locatello2020object}. The transformer encoder focuses on learning spatio-temporal features from action-observation sequences. Utilizing the slot attention mechanism, features are grouped at each temporal location, resulting in $K$ slot representations. The decoder subsequently decodes these slot representations to generate action logits. Notably, this parallelizable process enables faster training compared to existing benchmarks.

In multi-agent reinforcement learning, transformers have proven effective in modeling relations among agents and the environment~\citep{Zhang2022RelationalRV}. \citet{lilearning} proposed replacing \glspl{rnn} with a transformer encoder for robust temporal learning. Similarly, \citet{zhang2022tvenet} used a visual feature extractor based on the \gls{vit} architecture to obtain more robust representations for robotic visual exploration. Their network, utilizing self-attention, outperformed \gls{cnn} backbones in robotic tasks.

Transformers have been widely adopted in scenarios involving the processing of multimodal information. \citet{yang2022multi} introduced scene-fusion transformers that fuse observed trajectories and scene information to generate expressive representations for trajectory prediction. To reduce computational complexity, they employ sparse self-attention. \citet{zhang2022one} utilize transformers to integrate visual and text features effectively. 

\subsection{Enhancing Transferability and Generalization} 

An inherent difficulty faced by \gls{rl} is generalizing to new unseen tasks~\citep{levine2020offline}. This difficulty results from the intrinsic differences between various \gls{rl} tasks (e.g., autonomous driving and drug discovery). While meta-learning methods such as \gls{maml}~\citep{finn2017model} have been developed to generalize to new tasks with different distributions using limited data, these methods are hard to use in \gls{rl} due to poor sample-efficiency and unstable training~\citep{liu2019taming}. 

Transformers have shown great potential for meta-reinforcement learning (TrMRL), as demonstrated by \citet{Melo2022TransformersAM}. Transformers are excellent at handling long sequences and capturing dependencies over a long period, which enables them to adapt to new
tasks using self-attention quickly. In TrMRL, the proposed agent uses self-attention blocks to create an episodic memory representing a consensus of recent working memories. The transformer architecture encodes the working memory and tasks as a distribution over these memories. During meta-training, the agent learns to differentiate tasks and identify similarities in the embedding space. This approach performs comparably or better than PEARL~\citep{rakelly2019efficient} and \gls{maml}~\citep{finn2017model}. It is particularly efficient in memory refinement and task association. 

\citet{Shang2021StARformerTW} introduced the state-action-reward transformer (StARformer) to model multiple data distributions by learning transition representations between individual time steps of the state, action, and reward. StARformer consists of the step transformer and the sequence transformer. The step transformer uses self-attention to capture a local representation that understands the relationship between state-action-reward triplets within a single time-step window. The sequence transformer combines these local representations with global representation in the form of pure state features extracted as convolutional features, introducing a Markovian-like inductive bias. This bias helps reduce model capacity while effectively capturing long-term dependencies.

\section{Transition Function Learning}
\label{sec3:world}

\begin{figure}[h]
\includegraphics[scale=0.90]{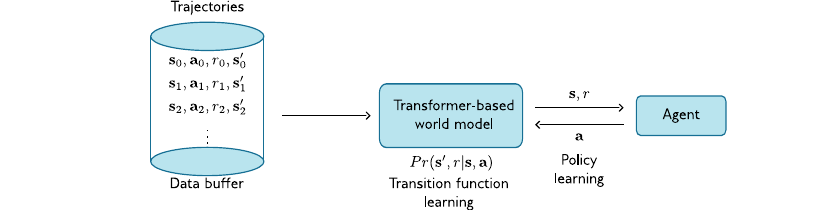}
\centering
\caption{A transformer can be employed to learn the dynamics of the simulator by modeling the transition function. This is accomplished by leveraging pre-collected trajectories of environment interaction. The transformer-based world model is trained using these trajectories. An \gls{rl} policy~$\pi$ is then trained to learn the task using this learned world model without requiring further environmental interaction.}
\label{fig:transition-function}
\end{figure}

The transition function~${p(\mathbf{s}', r|\mathbf{s}, \mathbf{a})}$ describes how the environment transitions from the current state~$\mathbf{s}$ to the next state~$\mathbf{s'}$ and issues rewards~$r$ in response to the actions~$\mathbf{a}$ taken by the agent. Learning this function (Fig.~\ref{fig:transition-function}) and subsequently exploiting it to train an \gls{rl} agent is known as model-based \gls{rl} (Sec.~\ref{sec:background:rl}). Model-based \gls{rl} offers a significant advantage compared to model-free \gls{rl} approaches~\citep{moerland2020model}; it allows the agent to plan future trajectories for each action, improving robustness and safety. Interactions with the external environment can be computationally expensive, particularly when relying on simulations that mimic the real world~\citep{featherstone2014rigid}. If we learn the transition function, these interactions can be reduced. 

A standard method in \gls{mbrl}~\citep{DBLP:journals/corr/abs-1803-10122} involves training an end-to-end world model to represent the environment's dynamics accurately. For instance, TransDreamer~\citep{Chen2022TransDreamerRL} trains a single model that learns visual representations and dynamics using the evidence lower bound loss~\citep{DBLP:journals/ftml/KingmaW19}. However, this approach can result in inaccuracies in the learned world model. 

The masked world model (MWM)~\citep{seo2022masked} addresses this by decoupling visual representation and dynamics learning. This framework utilizes an autoencoder with convolutional layers and \gls{vit} to learn visual representations. The autoencoder reconstructs pixels based on masked convolutional features. A latent dynamics model is learned by operating on the representations from the autoencoder. An auxiliary reward prediction objective is
introduced for the autoencoder to encode task-relevant information. Importantly, this approach outperforms the strong \gls{rnn}-based model, DreamerV2~\citep{hafner2020mastering} in terms of both sample efficiency and final performance on various robotic tasks.

Learning the dynamics of the world~$\mathbf{z}_{t+1} \sim p_G (\mathbf{z}_{t+1}| \mathbf{z}_{\leq t}, \mathbf{a}_{\leq t})$ has been formulated as a sequence modeling problem in \gls{iris}~\citep{micheli2022Transformers}. This approach takes advantage of the transformer's ability to process sequences of discrete tokens. \Gls{iris} uses a discrete autoencoder to construct a language of image tokens, while a transformer models the dynamics over these tokens. By simulating millions of trajectories accurately, \gls{iris} surpasses recent methods in the Atari 100k benchmark~\citep{DBLP:conf/ijcai/BellemareNVB15} in just two hours of real-time experience. 

Building upon the auto-regressive nature of transformer decoders, \citet{robine2022Transformer} introduces the transformer-based world model (TWM). Based on the \gls{tr-xl} architecture, TWM learns the transition function from real-world episodes while attending to latent states, actions, and rewards associated with each time step. By allowing direct access to previous states instead of viewing
them through a compressed recurrent state, the \gls{tr-xl} architecture enables the world model to learn long-term dependencies while maintaining computational efficiency.

\section{Reward Learning}
\label{sec3:reward} 

The reward function is crucial in \gls{rl} as it quantifies the desirability of different actions~$\mathbf{a}$ for a given state~$\mathbf{s}$, guiding the learning process. Typically, reward functions are predefined by human experts who carefully consider relevant factors based on their domain knowledge. However, designing an appropriate reward function is challenging in real-world scenarios, requiring a deep understanding of the problem domain. Moreover, manually designing it introduces bias and may lead to sub-optimal behavior. 

Recent research has explored different approaches for learning reward functions by integrating human data in various forms, such as real-time feedback, expert demonstrations, preferences, and language instructions. Transformers have proven valuable in these contexts. The transformer architecture is particularly advantageous with non-Markovian rewards, which are characterized by delays and dependence on the sequence of states encountered during an episode (e.g., when rewards are only provided at the end). Transformers efficiently capture dependencies across input sequences, making them well-suited to handle such scenarios.

\begin{figure}[h]
\includegraphics[scale=0.90]{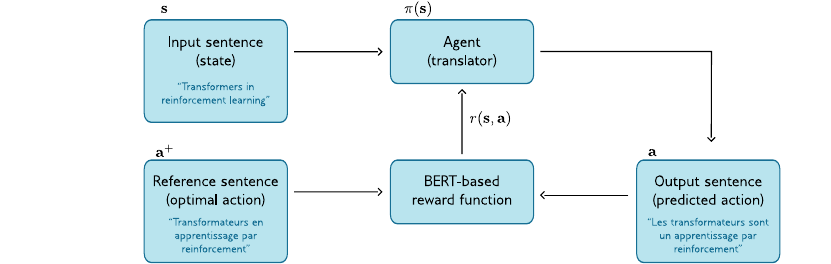}
\centering
\caption{\citet{nakatani2022comparing} use a \gls{bert}-based reward function~$r(\mathbf{s,a})$ to provide feedback to the \gls{rl} policy~$\pi$ to learn the task of machine translation. Given the input sentence~$\mathbf{s}$, the policy~$\pi$ translates it into a different language. This predicted action~$\mathbf{a}$ is then compared with the optimal action~$\mathbf{a^+}$ using \gls{bert}, which provides feedback to the agent in the form of a reward~$r$.}
\label{fig:reward-function}
\end{figure}

The preference transformer~\citep{kim2023preference} model captures human preferences by focusing on crucial events and modeling the temporal dependencies inherent in human decision-making processes; it effectively predicts non-Markovian rewards and assigns appropriate importance weights based on the trajectory segment. This approach reduces the effort required for designing reward functions and enables handling complex control tasks such as locomotion, navigation, and manipulation.

To train an \gls{rl} policy for generating text that aligns with human-labeled ground truth, the bilingual evaluation understudy (BLEU) score~\citep{papineni2002bleu} is often used as a reward function. However, BLEU may not consistently correlate strongly with human evaluation. In~\citep{nakatani2022comparing}, a \gls{bert}-based reward function is introduced, demonstrating a higher correlation with human evaluation. This approach leverages a pre-trained \gls{bert} model (Fig.~\ref{fig:reward-function}) to assess the semantic similarity between the generated and reference sentences and update the policy accordingly. 

\section{Policy Learning}
\label{sec3:policy}

Policy learning is central to \gls{rl}; it involves learning the policy $\pi(\mathbf{s})$ which the agent uses to select actions~$\mathbf{a}=\pi(\mathbf{s})$ with the objective of maximizing the discounted cumulative reward~$g$. 
Transformers have been used for modeling $\pi(\mathbf{s})$ in various scenarios, including off-policy, on-policy, and offline \gls{rl}.

\subsection{Offline RL with the Decision Transformer}

Offline \gls{rl} trains a policy using a limited, static dataset of previously collected experiences. This is different from online or off-policy \gls{rl} approaches (which continuously interact with the environment to update their policies) since the agent cannot collect experience beyond the fixed dataset, which limits its ability to learn, explore, and improve performance. 

\begin{figure}[h]
\includegraphics[scale=0.6]{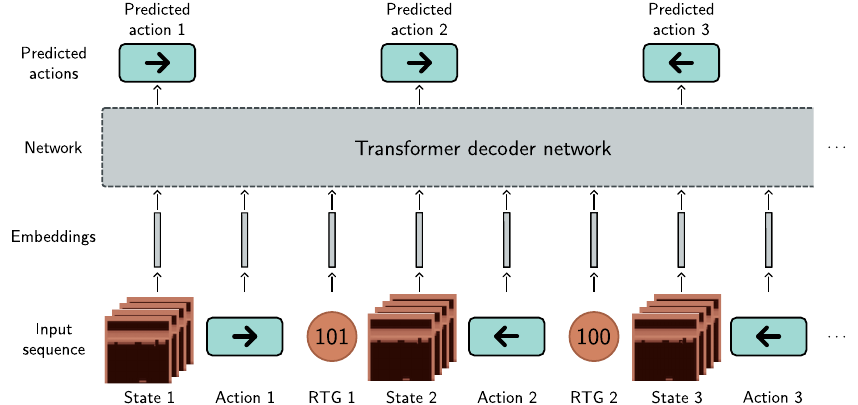}
\centering
\caption{The decision transformer approaches offline \gls{rl} as a sequence-prediction task. The network uses a sequence of states, actions, and return-to-go (RTG) to predict the next action. At inference, the RTG acts as a condition for the policy to generate a given set of actions.}
\label{DT}
\end{figure}

The \gls{dt}~\citep{chen2021decisiontransformer} (Fig.~\ref{DT}) is an offline \gls{rl} method that uses the upside-down \gls{rl} paradigm (see Sec.~\ref{sec:background:rl}). It uses a transformer-decoder to predict actions conditioned on past states, past actions, and expected return-to-go (the sum of the future rewards). The parameters are optimized by minimizing the cross-entropy (discrete) or mean square error (continuous) loss between the predicted and actual actions. 

\Gls{dt} uses the \gls{gpt} architecture to address the credit-assignment problem; the self-attention mechanism can associate rewards with the corresponding state-action pairs across long time intervals. This also allows the \gls{dt} policy to learn effectively even in the presence of distracting rewards~\citep{hung2019optimizing}. Empirical experiments demonstrate that the \gls{dt} outperforms state-of-the-art model-free offline approaches on offline datasets such as Atari and Key-to-Door tasks.

The \gls{dt} is a model-free approach that predicts actions based on past trajectories without forecasting new states, so it can't plan future actions. This limitation is addressed by the \gls{tt}~\citep{Janner2021ReinforcementLA}, an \gls{mbrl} approach that formulates \gls{rl} as a conditional sequence modeling problem. \Gls{tt} models past states, actions, and rewards to predict future actions, states, and rewards effectively. Using rewards as inputs prevent myopic behavior and enable the agent to plan future actions through search methods like beam-search~\citep{negrinho2018learning}.

This task-specific conditioning of agents offers flexibility in learning complex tasks. Prompt-based \gls{dt}~\citep{xu2022prompting}, enables few-shot adaptation in offline \gls{rl}. The input trajectory, which acts as a prompt, contains segments of few-shot demonstrations, encoding task-specific information to guide policy generation. This approach allows the agent to exploit offline trajectories collected from different tasks and adapt to new scenarios for generalizing to unseen tasks. Similarly, the text decision transformer (TDT)~\citep{Putterman2021PretrainingFL} employs natural language signals to guide policy-based language instruction in the Atari-Frostbite environment.

However, \glspl{dt} face several challenges. They struggle to learn effectively from sub-optimal trajectories. In a stochastic environment, their performance tends to degrade since the action taken may have been sub-optimal, and the achieved outcomes are merely a result of random environment transition. Insufficient distribution coverage of the environment is another challenge in offline \gls{rl} approaches like \gls{dt}. To overcome these challenges, solutions such as \gls{qdt}~\citep{yamagata2022q} re-label the return-to-go using a more accurate learned $Q$-function. \Gls{esper}~\citep{paster2022you} addresses stochastic performance degradation by conditioning on average return. Additionally, \gls{boot}~\citep{wang2022bootstrapped}, incorporates bootstrapping to generate more offline data. By adopting these approaches, the learning capabilities of \glspl{dt} can be improved, enabling more effective and robust policies in various scenarios. 

\subsection{Online RL with Transformers} 

Transformers have also been applied to online \gls{rl}, where the agent interacts with the environment while learning. In realistic environments, issues such as noisy sensors, occluded images, or unknown agents introduce the problem of partial observability. This makes it difficult for agents to choose the correct action (Sec.~\ref{sec:background:chall}). Here, retaining recent observations in memory is crucial to help disambiguate the true state. Traditionally, this problem has been approached using \glspl{rnn}, but transformers can provide better alternatives.

\begin{figure}[h]
\includegraphics[scale=0.90]{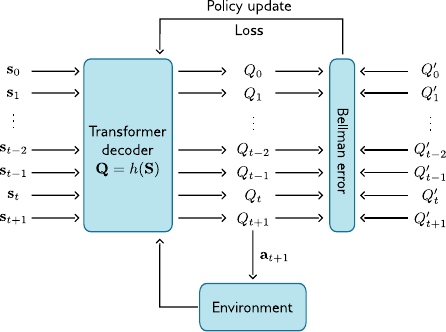}
\centering
\caption{The deep transformer $Q$-network~\citep{esslinger2022deep} employs a transformer decoder to learn a policy~$Q \sim \pi(\mathbf{S})$ that maps a given state to its corresponding $Q$ value for each action. During training, the network predicts $Q$ values for all the past observations, enabling policy updates via Bellman error, which measures the discrepancy between the current and updated value estimates obtained through the \gls{td} error. The network uses the current state’s $Q$ value to predict the optimal action at inference.}
\label{fig:policy-learning}
\end{figure}

The deep transformer $Q$-network (DTQN)~\citep{esslinger2022deep} addresses the challenge of partially observable environments using a transformer decoder architecture. At each time step of training, it receives the agent's previous $k$ observations and generates $k$ sets of $Q$-values. This unique training strategy encourages the network to predict $Q$-values even in contexts with incomplete information, leading to developing a more robust agent. During the evaluation, it selects the action with the highest $Q$-values from the last time-step in its history (Fig.~\ref{fig:policy-learning}).

The DTQN incorporates a learned positional encoding, which enables the network to adapt to different domains by learning domain-specific temporal dependencies. This domain-specific encoding matches the temporal dependencies of each environment and allows the DTQN to adapt to environments with varying levels of temporal sensitivity. The DTQN demonstrates superior learning speed and outperforms previous recurrent approaches in various partially observable domains, including gym-gridverse, car flag, and memory cards~\citep{DBLP:journals/corr/abs-2303-01859}. 

\subsection{Transformers for Multi-Agent Reinforcement Learning} 

\Gls{marl} (Sec.~\ref{sec:background:rl}) presents unique challenges as agents learn and adapt their behaviors through interactions with other agents and the environment. One such challenge stems from the model architecture's fixed input and output dimensions, which means that different tasks must be trained independently from scratch~\citep{shao2018starcraft, wang2020few}. Consequently, zero-shot transfer across tasks is limited. 

Another challenge arises from the failure to disentangle observations from different agents~\citep{hu2021updet}. When all information from various agents or environments is treated equally, it can result in misguided decisions by individual agents. This challenge becomes particularly prominent when utilizing a centralized value function, which serves as a shared estimate of actions and state value across multiple agents to guide their behavior~\citep{chen2023credit}. As a result, appropriately assigning credit to individual agents becomes difficult.

\begin{figure}[h]
\includegraphics[scale=0.90]{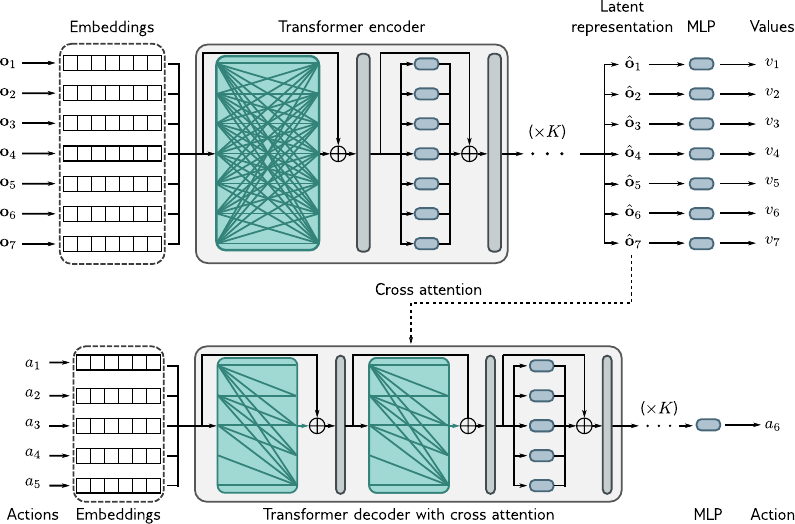}
\centering
\caption{The multi-agent transformer~(MAT)~\citep{wen2022multi} employs an encoder-decoder architecture. The encoder receives a series of observations from the agents and converts them into latent representations. These representations are then fed into the decoder. The decoder generates each agent’s optimal action step-by-step
sequentially and auto-regressively. To ensure proper training, the masked attention blocks restrict agents to only access actions from preceding agents.}
\label{fig:multi-agent}
\end{figure}

The universal policy decoupling transformer (UPDeT)~\citep{hu2021updet} is designed to handle challenges in tasks with varying observation and action configuration requirements. It achieves this by separating the action space into multiple action groups, effectively matching related observations with corresponding action groups. UPDeT improves the decision-making process by employing a self-attention mechanism and optimizing the policy at the action-group level. This enhances the explainability of decision-making while allowing for high transfer capability to new tasks.

This characteristic is also observed in the multi-agent transformer (MAT)~\citep{wen2022multi}. MAT (Fig.~\ref{fig:multi-agent}) transforms the joint policy search problem into a sequential decision-making process, allowing for parallel learning of agents' policies regardless of the number of agents involved. The encoder utilizes the self-attention mechanism to process a sequence of each agent's observations, capturing their interactions. This generates a sequence of latent representations that are then fed into the decoder. The decoder, in turn, produces each agent's optimal action in an auto-regressive and sequential manner. As a result, MAT possesses robust generalization capabilities, surpassing \gls{mappo}~\citep{lohse2021implementing}, and \gls{happo}~\citep{kuba2021trust} in few-shot experiments on multi-agent MuJoCo tasks. 

However, \gls{marl} faces limitations in real-world applications due to the {\em curse of many agents}~\citep{wang2020breaking}, which stems from the exponentially growing state-action space as the number of agents increases. This presents challenges in learning the value functions and policies of the agents, leading to
inefficient relational reasoning among them and credit-assignment problems. Concatenating the state-action spaces of individual agents and treating them as a single-agent problem leads to exponential state and time complexity~\citep{zhou2020learning}. Additionally, independent learning of policies may struggle to converge without cooperation~\citep{gupta2021hammer}.

TransMix~\citep{junaid2022Transformer} tackles the challenge through a centralized learning approach, enabling agents to exchange information during training. During policy execution, each agent relies on a partially observable map. The action space in the star-craft multi-agent challenge (SMAC)~\citep{vinyals2017starcraft} encompasses various actions, including moving units, attacking enemies, gathering resources, constructing buildings, and issuing commands to control the game state. Utilizing transformers, TransMix captures global and local contextual interactions among agent $Q$-values, histories, and global state information, facilitating efficient credit assignment.

The transformer's ability to reason about relationships between agents improves results in both model-free \gls{marl} (regardless of the number of agents) and model-based \gls{marl} (with a logarithmic dependence on the number of agents)~\citep{guedj2019primer}. Notably modeling the transformer's self-attention with other neural network types requires an impractically large number of trainable parameters, highlighting the significance of self-attention in capturing agent interactions~\citep{guedj2019primer}. Moreover, the transformer's performance remains stable across different agent counts, with accuracy impacted by neural network depth~\citep{guedj2019primer}, making it highly efficient for \gls{marl}.

\section{Training Strategy}
\label{sec:trainstra}
Training transformers poses challenges due to their reliance on residual branches, which amplify
minor parameter perturbations, disrupting model output~\citep{liu2020understanding}; specialized optimizers and weight initializers are needed for successful training. Likewise, the training of \gls{rl} policies can be unstable~\citep{nikishin2018improving} and require distinct strategies for achieving optimal performance. Hence, the integration of transformers into \gls{rl} is particularly challenging. These challenges can manifest as sudden or extreme changes in performance during training, impeding effective learning and generalization.

The standard transformer architecture is difficult to optimize using \gls{rl} objectives and needs extensive hyper-parameter tuning, which is time-consuming. Here, we review strategies for training transformers in \gls{rl}. These include pre-training and transfer learning to expedite learning, improved weight initialization to mitigate gradient issues, and efficient layer utilization for capturing relevant information.

\subsection{Pre-Training and Transfer Learning}
Transformers can be pre-trained on large, reward-free datasets, providing opportunities to fine-tune when only small annotated datasets are available. \citet{Meng2021OfflinePM} propose using \gls{dt} to pre-train agents on large, reward-free offline datasets of prior interactions. During pre-training, reward tokens are masked, allowing the transformers to learn to predict actions based on the previous state and action content while extracting behavior from the dataset. This pre-trained model can then be fine-tuned with a small, reward-annotated dataset to learn the skills necessary to achieve the desired behavior based on the reward function.

Transfer learning is challenging when the environment dynamics change. A training method for \gls{dt}~\citep{Boustati2021TransferLW} addresses this challenge by using counterfactual reasoning. It generates counterfactual trajectories in an alternative environment, which are used to train a more adaptable learning agent. This process aids in regularizing the agent's internal representation of the environment, enhancing its adaptability to structural changes. Moreover, unsupervised pre-training of vision and sequence encoders has also improved downstream few-shot learning performance~\citep{Putterman2021PretrainingFL}. By leveraging pre-trained models, the agent can quickly adapt to new, unseen environments and achieve higher performance with limited training data. 

\subsection{Stabilizing Training}
In the \gls{rl} setting, transformer models require learning rate warmup to prevent divergence caused by backpropagation through the layer normalization modules, which can destabilize optimization. To enhance stability, \citet{Melo2022TransformersAM} proposes to use T-Fixup initialization~\citep{pmlr-v119-huang20f}. This applies Xavier initialization~\citep{DBLP:journals/jmlr/GlorotB10} to all parameters except input embeddings, eliminating the need for learning rate warmup and layer normalization. It is crucial in environments where learned behavior guides exploration; it addresses instability during early training stages when policies are more exploratory and prevent convergence to sub-optimal policies.

The \gls{gtr-xl} architecture~\citep{parisotto2020stabilizing} has demonstrated promising results in stabilizing \gls{rl} training and improving performance. It improves upon the original \gls{tr-xl} architecture by applying layer normalization exclusively to the input stream within the residual model rather than the shortcut stream. This modification allows the initial input to propagate through all the layers, promoting training stability. \Gls{gtr-xl} replaces the residual connection with a \gls{gru}-style gating mechanism. This gating mechanism regulates information flow through the network controlling the amount of information passed via the shortcut. This added flexibility enhances the model's adaptability to \gls{rl} scenarios and facilitates stable training.

\section{Interpretability}
\label{sec:explainab}
Interpretability of the learned \gls{rl} policies is desirable in safety-critical applications like healthcare and autonomous driving~\citep{glanois2021survey}. This helps in building trust, facilitating debugging, and promoting ethical and fair decision-making. However, achieving interpretability has been a significant challenge and a bottleneck in the progress of \gls{rl}~\citep{milani2022survey, heuillet2021explainability}.

One way to interpret transformers is to visualize the attention weights using heatmaps~\citep{zhang2022exploiting}. This helps to understand which features are used to learn the particular task. In multi-agent scenarios, these visualizations reveal the localized areas of the input space where the individual agents focus their attention, facilitating coordinated and cooperative behavior. For instance, \citet{motokawa2021mat} introduce a multi-agent transformer deep $Q$-network (MAT-DQN) that integrates transformers into a deep $Q$-network. Using heatmaps, MAT-DQN provides insights into the important input information that influences the agent's decision-making process for cooperative behavior. 

Analyzing attention heatmaps unveils the agent's ability to consider other agents, relevant objects, and pertinent tasks, allowing for a clear interpretation of the policy. Such visualization is critical in sparse reward settings, where understanding which past state had the most influence on decision-making is crucial. Attention-augmented memory (AAM)~\citep{qu2022interpretable} exemplifies this by combining the current observation with memory. This enables the agent to understand ``what'' the agent observes in the current environment and ``where'' it directs its attention in its memory.

An interesting method for enhancing interpretability involves the use of transformers in neuro-symbolic policies~\citep{bastani2022interpretable}. Neuro-symbolic policies combine programs and neural networks to improve interpretability and flexibility in \gls{rl} tasks. Specifically, a neuro-symbolic transformer is a variant of the traditional transformer model that incorporates programmatic policies into the attention mechanism. Instead of utilizing a neural network, the attention layer employs a program to determine the relevant inputs to focus on. These programmatic policies can take various forms, including decision trees, rule lists, and state machines. This approach improves interpretability by providing a more precise understanding and visualization of why agents attend to specific inputs. 

However, it has been demonstrated that attention weights alone are unreliable predictors of the importance of intermediate components in \gls{nlp}~\citep{serrano2019attention, bai2021attentions}, leading to inaccurate explanations of model decisions; learned attention weights often highlight less meaningful tokens and exhibit minimal correlation with other feature importance indicators like gradient-based measures. Furthermore, relying solely on attention weights can result in fragmented explanations that overlook most other computations. Recent work has introduced the assignment of a local relevancy score~\citep{chefer2021transformer}. These scores are propagated through layers to achieve class-based separation and enhance the interpretability of the transformers. This approach holds promise for future research to improve the interpretability of \gls{rl} policies. 


\section{Applications}
\label{sec:applic}
\Gls{rl} has traditionally been constrained to unrealistic scenarios in virtual environments. However, with modern deep neural network architectures, there has been a notable shift towards employing \gls{rl} to address a broader range of practical challenges. The following section describes real-world applications where \gls{rl} powered by transformers can make a substantial impact.

\subsection{Robotics}
\label{sec4:robotics}
In robotics, autonomous agents automate complex real-world tasks; a classic example is autonomous driving. Here, learning the \gls{rl} policy for trajectory planning is essential: it involves forecasting the future positions of one or more agents in an environment while considering contextual information. This requires adequate planning and coordination among agents by modeling their spatial and temporal interactions. 

Several studies have proposed to use transformers for processing sequences of high-dimensional scene observations for predicting actions. A recent study~\citep{kargar2022vision} uses \gls{vit} to extract spatial representations from a birds-eye view of the ego vehicle to learn driving policies. Compared with \glspl{cnn}, \glspl{vit} are more effective in capturing the global context of the scene. The attention mechanism used in \glspl{vit} allow the policy to discern the neighboring cars that are pivotal in the decision-making process of the ego vehicle. As a result, the \gls{vit}-based DQN agent outperforms its \gls{cnn}-based counterparts. \citet{liu2022augmenting} introduce a transformer architecture to encode heterogeneous information, including the historical state of the ego vehicle and candidate route waypoints, into the scene representation. This approach enhances sample efficiency and results in more diverse and successful driving behaviors during inference.
The object memory transformer~\citep{fukushima2022object} explores how long-term histories and first-person views can enhance navigation performance in object navigation tasks. An object scene memory stores long-term scene and object semantics, focusing attention on the most salient event in past observations. The results indicate that incorporating long-term object histories with temporal encoding significantly enhances prediction performance.

Transformers also excel in capturing both spatial relationships and intra-agent interactions,
making them ideal for facilitating cooperative exploration and developing intelligent embodied
agents. \Gls{maans}~\citep{Yu2021LearningEM} addresses the challenge of cooperative multi-agent exploration~\citep{oroojlooy2022review}, where multiple agents collaborate to explore unknown spatial regions. This approach extends the single-agent active neural SLAM~\citep{chaplot2020learning} method to the multi-agent setting and utilizes a multi-agent spatial planner with a self-attention-based architecture known as the Spatial-TeamFormer. This hierarchically integrates intra-agent interactions and
spatial relationships, employing two layers: An individual spatial encoder that captures spatial features for each agent, and a team relational encoder for reasoning about interactions among agents. To focus on spatial information, the intra-agent self-attention performs spatial self-attention over each agent's spatial map independently. The team relation encoder focuses on capturing team-wise interactions without leveraging spatial information. This allows \gls{maans} to outperform planning-based competitors in a photo-realistic environment, as shown in experiments on Habitat~\citep{savva2019habitat}.

\subsection{Medicine}
\label{sec4:medicine}
\Gls{rl} has the potential to assist clinicians; tasks involving diagnosis, report generation, and drug discovery can be considered sequential decision-making problems~\citep{Yu2019ReinforcementLI}. 
\newline

\noindent \textbf{Disease Diagnosis.} Diagnosing a medical condition involves modeling a patient's information (e.g., treatment history, present signs, and symptoms) to accurately understand the disease. \citet{chen2022dxformer} propose a model for disease diagnosis called the DxFormer. This employs a decoder-encoder transformer architecture, where the decoder inquires about implicit symptoms. At the same time, the encoder is responsible for disease diagnosis, which models the input sequence of symptoms as a sequence classification task. To facilitate symptom inquiry, the decoder is formulated as an agent that interacts with a patient simulator in a serialized manner, generating possible symptom tokens that may co-occur with prior known symptoms and inquiring about them. The inquiry process proceeds until the confidence level in the predicted disease surpasses a selected threshold, thus enabling a more accurate and reliable diagnosis.\newline

\noindent \textbf{Clinical Report Generation.} \Gls{rl} can generate medical reports from images by employing appropriate evaluation metrics such as human evaluations or consensus-based image description evaluation (CIDEr)~\citep{vedantam2015cider} and BLEU metrics as rewards. Previous approaches to medical image captioning were constrained by their reliance on \glspl{rnn} for text generation, which often resulted in slow performance and incoherent reports, as highlighted by \citet{Xiong2019ReinforcedTF}. To address this limitation, their work introduces an \gls{rl} approach based on transformers for medical image captioning. Initially, a pre-trained \gls{cnn} is employed to identify the region of interest in chest X-ray images. Then, a transformer encoder is utilized to extract the visual features from the identified regions. These features serve as input to the decoder, which generates sentences describing the X-ray scans. Similarly~\citet{miura2021improving} used a meshed-memory transformer (M2 Trans)~\citep{cornia2020meshed} that generates radiology reports, proving more effective than traditional \gls{rnn} and transformer models. M2 Trans incorporates a \gls{cnn} to extract image regions. These regions are then encoded using a memory-augmented attention process. This involves assigning attention weights to the image based on prior knowledge stored in memory matrices that capture relationships between different regions. This model is trained using
rewards, aiming to enhance generated reports' factual completeness and consistency.
\newline

\noindent \textbf{Drug Discovery.} \Gls{rl} has the potential to accelerate drug discovery efforts. It has been utilized to bias or fine-tune generative models, enabling the generation of molecules with more desirable characteristics, such as bioactivity. Traditional generative models for molecules, such as \glspl{rnn} or \glspl{gan}~\citep{goodfellow2020generative} have limitations in satisfying specific constraints, such as synthesizability or desirable physical properties. Recent research~\citep{Wang2021MulticonstraintMG, liTransformergan, Yang2021TransformerBasedGM, Liu2021DrugExVS} uses transformers as generative models for molecular generation. These approaches generate better plausible molecules with rich semantic features. A discriminator grants rewards that guide the policy update of the generator. These works demonstrate that transformer-based methods significantly improve capturing and utilizing structure-property relations, leading to higher structural diversity and a broader range of scaffold types for the generated molecules.

\subsection{Language Modeling}
\label{sec4:tav}

Language modeling involves understanding the sequential context of language to perform diverse tasks like recognition, interpretation, or retrieval. Large language models like \gls{gpt} leverage pre-training on vast corpora, enabling them to generate fluent, natural language by sampling from the learned distribution, thus minimizing the need for extensive domain-specific knowledge engineering. However, these models face challenges in maintaining task coherence and goal-directedness. \citet{alabdulkarim2021goal} use \gls{ppo} to fine-tune an existing transformer-based language model specifically
for story generation to address this issue. This approach inputs a text prompt and generates a story based on
the provided goal. This policy is updated using a reward mechanism that considers the proximity of the generated story to the desired input goal and the frequency of verb occurrence in the story compared to the goal.

Several studies use additive learning to benefit from pre-trained language models with limited data, incorporating a task-specific adapter over the frozen pre-trained language model. \citet{Jo2022SelectiveTG} use \gls{rl} to selectively sample tokens between the general pre-trained language model and the task-specific adapter. The authors argue that this enables the adapter to focus solely on the task-relevant component of the output sequence, making the model more robust to over-fitting. \citet{cohen2022dynamic} introduce a conversational bot powered by \gls{rl} where pre-trained models encode conversation history. Given that the action space for dialogue systems can be very large, the authors propose limiting the action space to a small set of generated candidate actions at each conversation turn. They use $Q$-Learning-based \gls{rl} to allow a dynamic action space at each stage of the conversation. 

Increasing the size of language models alone does not necessarily mitigate the risk of toxic biases in the training data. Several RL-based approaches have been proposed to better align these models with the user's intended objectives. To align GPT-3 to the user’s preferred intentions, \citet{ouyang2022training} introduce InstructGPT. First, the authors propose to collect a set of human-written demonstrations of desired output behavior and fine-tune GPT-3 with supervised learning. Next, a reward model is trained on model outputs ranked from best to worst. Using this reward model, the model is further optimized with RL using \gls{ppo}. Results demonstrate that InstructGPT, with 1.3B parameters, produces preferable outputs than much larger models such as GPT-3 with 175B parameters. \citet{faal2022reward} propose an alternative method to mitigate toxicity in language models via fine-tuning with \gls{ppo}. They use a reward model based on multi-task learning to mitigate unintended bias in toxicity prediction related to various social identities.

\subsection{Edge and Cloud Computing}
\label{sec4:edgecloud}
\Gls{rl} is a valuable tool for optimizing the performance of decision-making systems that require real-time adaptation to changing conditions, such as those used in edge and cloud computing. In edge computing, \gls{rl} can optimize resource-constrained \gls{iot} devices' performance~\citep{chen2021deep}. In cloud computing, \gls{rl} can be used to optimize resource allocation and scheduling in large-scale distributed systems~\citep{gondhi2017survey}. Integrating transformers with \gls{rl} in these two settings can be particularly useful as they can handle high-dimensional sensory states~\citep{ho2019axial} and sequences of symbolic states~\citep{bhattamishra2020computational}. 

A distributed deep \gls{rl} algorithm proposed by~\citet{wang2022distributed} utilizes transformers to model the policy for optimizing offloading strategies in vehicular networks. These networks enable vehicle-to-vehicle communication. To represent the input sub-task priorities and dependencies, a \gls{dag} is used. Thereafter the attention mechanism employed by the transformer allows for the efficient extraction of state information from this \gls{dag}-based topology representation. This facilitates informed offloading decisions. The reward function used in this algorithm optimizes for latency and energy consumption providing valuable feedback. This approach enables faster convergence of the vehicular agent to equilibrium.

\subsection{Combinatorial Optimization}
\label{sec4:combopt}
Combinatorial optimization involves finding the values of a set of discrete parameters that minimize cost functions~\citep{mazyavkina2021reinforcement}. Recently, transformer-based models have shown promise in combinatorial optimization (e.g., for the traveling salesman and routing problems) due to their ability to handle sequential data and model complex relationships between entities. \newline

\noindent \textbf{Travelling Salesman.} This problem is a classical combinatorial optimization problem commonly found in crew scheduling applications. It has been formulated as an \gls{rl} problem by~\citet{smith2022attention}. The problem is NP-hard and not NP-complete. The high polynomial complexities of brute-force algorithms necessitate the development of faster methods. This study uses a \glsfirst{dt} to solve this challenge. The \gls{dt} is fed random walks as input and aims to find the optimal path among all nodes. The \gls{dt} has the advantage of scaling with pseudo-linear time, as it only needs to predict once per node in the route. This significantly improves over previous methods, such as dynamic programming and simulated annealing, which have polynomial and exponential complexity. However, the \gls{dt} could not always accurately model the travelling salesman problem leading to inconsistent performance. \newline

\noindent \textbf{Routing.} Identifying the most efficient route between two nodes in a graph is important in industries such as transportation, logistics, and networking~\citep{mor2022vehicle}. Traditional heuristic-based algorithms may not always yield the optimal solution, as adapting to changing conditions is challenging~\citep{wu2021learning}. \Glspl{gnn} have been used to address these challenges~\citep{lu2020learning}. However, these may not be sufficient for handling data with complex inter-relationships and structures, motivating the use of transformers. In addition, routing often requires optimizing for multiple constraints, such as cost, time, or distance, which can be tackled using \gls{rl}. A transformer-based policy has been proposed by~\citet{wang2023deep} to tackle the routing problem using a standard transformer encoder with positional encoding, which ensures translation invariance of the input nodes. A \gls{gnn} layer is used in the decoder, enabling consideration of the graph's topological structure formed by the node relationships. The policy is then trained using the REINFORCE algorithm~\citep{10.1007/BF00992696}. This approach improves learning efficiency and optimization accuracy compared to traditional methods while providing better generalization in new scenarios.

\subsection{Environmental Sciences}
\label{sec4:clch}
\Gls{rl} algorithms can help address climate change by optimizing the behavior of systems and technologies for reducing greenhouse gas emissions and mitigating the impacts of climate change~\citep{strnad2019deep}. These algorithms can learn and adapt to multiple constraints, optimizing performance without compromising productivity. However, in such settings, \gls{rl} algorithms must rely on past context stored in memory, and integrating prior knowledge is crucial for their success, suggesting the use of transformers.

\citet{nasir2022deep} formulate the problem of closed-loop reservoir management problem as a \gls{pomdp}. In subsurface flow settings, such as oil reservoirs, the goal is to extract as much oil as possible while minimizing costs and environmental impact. However, this requires making decisions about well pressure settings often complicated by geological model uncertainties. This work models the \gls{rl} policy with \gls{ppo} using temporal convolution and gated transformer blocks for an efficient and effective representation of the state. The framework's training is accomplished with data generated from flow
simulations across an ensemble of prior geological models. After appropriate training, the policy instantaneously maps flow data observed at wells to optimal pressure. This approach helps reduce computational costs and improve decision-making in subsurface flow settings.

\citet{wang2022stabilizing} introduce transformer-based multi-agent actor-critic framework (T-MAAC) leveraging \gls{marl} algorithms to stabilize voltage in power distribution networks. This framework recognizes the need for coordination among multiple units in the grid to handle the rapid changes in power systems resulting from the increased integration of renewable energy and tackles this problem by using \gls{marl} algorithms. The proposed approach introduces a transformer-based actor that takes the grid state representation as input and outputs the maximum reactive power ratio that each agent in the power distribution can generate. Subsequently, the critic approximates the global $Q$-values using the self-attention mechanism to model the correlation between agents across the entire grid. The policy is reinforced through feedback in the form of reward, aiming to control voltage within a safe range while minimizing power loss in the distribution network. This approach consistently enhances the effectiveness of the active voltage control task.

\subsection{Scheduling}
\label{sec4:schedul}

The scheduling problem involves determining the optimal arrangement of tasks or events within a specified time frame while considering constraints such as resource availability or dependencies between tasks~\citep{allahverdi2016survey}. This problem can arise in various contexts, such as scheduling jobs in manufacturing or optimizing computer resource usage, and can be approached using various techniques~\citep{parmentier2021learning}. Transformers are now being used to solve scheduling problems.

The job-shop scheduling problem (JSSP) is a classical NP-hard problem that involves scheduling a set of jobs on a set of machines, where each job has to be processed on each machine exactly once, subject to various constraints. An \gls{rl} approach for solving the JSSP using the \gls{dgerd} transformer is proposed by~\citet{chen2022deep}. This work uses the attention
mechanism and disjunctive graph embedding to model the JSSP, which allows complex
relationships between jobs and machines to be captured. In the context of JSSP, the attention module learns to prioritize certain jobs or machines based on their importance or availability. By doing
so, it can generate more efficient and robust schedules. The disjunctive graph embedding converts the JSSP instance into a graph representation to capture the structural properties, enabling better generalization and reducing over-fitting. This acts as an input for the \gls{dgerd} transformer consisting of a parallel-computing encoder and a recurrent-computing decoder. The encoder takes the disjunctive graph embedding of the JSSP instance and generates a set of hidden representations that capture the relevant features of the input. This hidden representation is then fed to the decoder to generate an output schedule sequentially. The policy is optimized using feedback from the environment in the form of makespan (length of time that elapses from the start of work to the end) and tardiness penalties. This helps in generating schedules that are both fast and reliable.

\subsection{Trading}
\label{sec4:trad}
Stock portfolio optimization involves choosing the optimal combination of assets to obtain the highest possible returns while minimizing risk~\citep{hieu2020deep}. This process can be challenging due to the various factors that can impact a portfolio's performance~\citep{haugh2001computational}, which include market conditions, economic events, and changes in the value of individual stocks. Various techniques can be used to optimize a portfolio, including modern portfolio theory and optimization algorithms~\citep{thakkar2021comprehensive}, and \gls{rl} is one such approach to automate the trading process. For this application, \gls{rl} involves training a model to make trading decisions based on historical data and market conditions to maximize the portfolio's return over time.

Although past performance may not indicate future results, data-driven approaches rely on past features to model a particular stock's expected future performance. This is because various historical data points, such as price trends, trading volumes, and market sentiment, may hint at a stock's future performance~\citep{milosevic2016equity}. In portfolio optimization, it is necessary to consider both short-term and long-term trends~\citep{ta2020portfolio}. The transformer architecture is well-suited for this task.

The first application of transformers in portfolio selection, as introduced by~\citet{xu2021relation}, involved using the \gls{rat}. This uses an encoder-decoder transformer architecture to model the \gls{rl} policy. The encoder takes in the sequential price series of assets, such as stocks and cryptocurrencies, as the input state. It performs sequential feature extraction, comprising a sequential attention layer for capturing patterns in asset prices and a relation attention layer for capturing correlations among assets. The decoder has a network resembling an encoder, with an additional decision-making layer that incorporates leverage and enables accurate decisions, including short sales, for each asset. The final action is determined by combining the initial portfolio vector, the short sale vector, and the reinvestment vector. The agent then receives reward-based feedback, measured as the log return of portfolios. To evaluate the proposed method, real-world cryptocurrency, and stock datasets are used and compared against state-of-art portfolio selection methods. The results demonstrate a significant improvement over existing approaches.

\subsection{Hyper-Parameter Optimization}
\label{sec4:hpo}
\Gls{hpo} involves finding the optimal set of hyper-parameters for training a machine learning model. Some commonly used hyper-parameters in machine learning models include the learning rate, batch size, number of hidden units in a neural network, and activation function. As such, finding the best combination of these hyper-parameters can be challenging for large models due to their correspondingly large search space~\citep{ali2023hyperparameter}. Manually setting hyper-parameter values is fast but requires expertise and domain knowledge~\citep{shawki2021automating}. Automated techniques like random search, grid search, or Bayesian optimization can automatically find ideal hyper-parameter combinations~\citep{bergstra2012random, snoek2015scalable}, but minimizing overall computational costs remains a challenge. Such auto-tuners rarely perform well for complex tasks and are prone to errors with increased model complexity~\citep{shawki2021automating}. 

\Gls{ame}~\citep{xu2022ame}, is a transformer-based search algorithm to enhance the selection of hyper-parameters that tackles these challenges. \Gls{ame} utilizes \gls{rl} and addresses \gls{hpo} without relying on distribution assumptions. The agent, or the searcher, is modeled using \gls{gtr-xl} and learns a series of state-to-action mappings based on rewards. In this context, the state refers to the combination of evaluated configurations, while an action corresponds to the new configuration chosen by the agent from the search space. Utilizing \gls{gtr-xl} improves the ability to capture relationships among different configurations through memory mechanisms and multi-head attention, thereby enabling attentive sampling. The agent is trained using feedback in the form of rewards, which promotes the generation of high-performance configurations and penalizes those leading to reduced performance. Consequently, it 
 effectively locates high-performance configurations within vast search space. Results demonstrate that the \gls{ame} algorithm surpasses other \glspl{hpo} like Bayesian optimization, evolutionary algorithms, and random search methods in terms of adaptability to diverse tasks, efficiency, and stability.

\section{Limitations}
\label{sec:limit}
As discussed above, transformers are gradually being integrated into \gls{rl} for various applications. Despite these advances, some limitations impede their widespread use. This section details these limitations and provides insights for future research. \newline

\noindent \textbf{Balancing Local \& Global Context.} In \gls{rl}, global contextual information is required for efficient high-level planning~\citep{barto2003recent}. This information is combined with additional nearby details, known as the local context, to predict low-level actions precisely. As detailed by~\citet{li2019enhancing, wang2021local}, transformers may not be as effective as other models in capturing local context. This limitation is mainly because of the self-attention mechanism, which compares queries and keys for all elements in a sequence using the dot product. This point-wise comparison does not directly consider the local context for each sequence position, which may lead to confusion due to noisy local points. Recent studies~\citep{lin2023batformer, wang2022uformer, wang2021boundary} inspired by \glspl{cnn} have proposed modifications to the original attention mechanism to balance the local and global context more effectively. These approaches include local window-based boundary-aware attention, allowing the model to focus on a small window of nearby details and the global context when making predictions. \newline

\noindent \textbf{Weak Inductive Bias.} \Glspl{cnn} and \gls{lstm}~\citep{hochreiter1997long} models have a strong inductive bias toward the dataset's structure, which helps narrow the search space and leads to faster training~\citep{d2021convit}. This makes them better suited for situations with less training data. However, transformers have a relatively weak inductive bias, making them more capable of finding general solutions~\citep{hessel2019inductive} but more susceptible to over-fitting, especially when less data is available. This limitation can be a significant challenge in \gls{rl}, where training a policy already requires millions of trajectories. Furthermore, learning models like the decision transformer require collecting trajectories from learned policies which can be even more challenging. One approach to counter weaker inductive bias in transformers is to use foundation models~\citep{zhou2023comprehensive, moor2023foundation}. Foundation models are pre-trained on large and diverse datasets, which allows them to learn general patterns that can be applied to a wide range of downstream tasks. The foundation model can achieve state-of-the-art results with less data by fine-tuning the pre-trained model on a smaller task-specific dataset. \newline

\noindent \textbf{Quadratic Complexity.} The self-attention mechanism of transformers becomes more computationally expensive as input sequence length increases due to the quadratic increase in pairwise comparisons between tokens~\citep{keles2023computational}. This limitation, along with hardware and model size constraints, restricts the ability of transformers to process longer input sequences, making them unsuitable for specific tasks that require substantial amounts of contextual information, like document summarization or genome fragment classification. This limitation can also pose challenges in \gls{rl} for applications requiring extended temporal modeling. However, recent works~\citep{katharopoulos2020transformers, lu2021soft, ren2021combiner} have provided methods to reduce this cost to linear or sub-quadratic, providing new possibilities for using transformers in applications that require longer input sequences.

\section{Conclusion}
This survey explored the diverse uses of transformers in \gls{rl}, including representation learning, reward modeling, transition function modeling, and policy learning. While the original transformer architecture has limitations, it can be modified for many \gls{rl} applications. We showcased the advances in transformers that have broadened the scope of \gls{rl} to real-world problems in robotics, drug discovery, stock trading, and cloud computing. Finally, we discussed the current limitations of transformers in \gls{rl} and ongoing research in this field. Given its versatility in addressing challenges such as partial observability, credit assignment, interpretability, and unstable training --- issues commonly encountered in traditional \gls{rl} --- we anticipate that the transformer architecture will continue to gain popularity in the \gls{rl} domain. \newline

\noindent \textbf{Acknowledgement.} We thank CIFAR, Google, CMLabs for funding the project, and Vincent Michalski for the valuable feedback.

\bibliographystyle{ACM-Reference-Format}
\bibliography{references}

\end{document}